\documentclass[10pt,letterpaper]{article}

\usepackage{ccn}
\usepackage{pslatex}
\usepackage{apacite}
\usepackage{graphicx}
\usepackage{amsmath}
\title{3D View Prediction Models of the Dorsal Visual Stream}
 
\author{{\large \bf Gabriel H. Sarch (gsarch@andrew.cmu.edu)} \\
  Neuroscience Institute, Carnegie Mellon University, Pittsburgh, PA, USA.
  \AND {\large \bf Hsiao-Yu Fish Tung (hytung@mit.edu)} \\
  Department of Brain and Cognitive Sciences, Massachusetts Institute of
Technology, Cambridge, MA, USA.
  \AND {\large \bf Aria Y. Wang (ariawang@cmu.edu)} \\
  Neuroscience Institute, Carnegie Mellon University, Pittsburgh, PA, USA.
  \AND {\large \bf Jacob S. Prince (jacob.samuel.prince@gmail.com)} \\
  Department of Psychology, Harvard University, Cambridge, MA, USA.
  \AND {\large \bf Michael J. Tarr (michaeltarr@cmu.edu)} \\
  Department of Psychology \& Neuroscience Institute, Carnegie Mellon University, Pittsburgh, PA, USA.
  }

\begin{document}

\maketitle

\newpage
\vspace{-.2 cm}
\section{Abstract}
{
\bf

\vspace{-.2 cm}
Deep neural network representations align well with brain activity in the ventral visual stream. However, the primate visual system has a distinct dorsal processing stream with different functional properties. To test if a model trained to perceive 3D scene geometry aligns better with neural responses in dorsal visual areas, we trained a self-supervised geometry-aware recurrent neural network (GRNN) to predict novel camera views using a 3D feature memory. We compared GRNN to self-supervised baseline models that have been shown to align well with ventral regions using the large-scale fMRI Natural Scenes Dataset (NSD). We found that while the baseline models accounted better for ventral brain regions, GRNN accounted for a greater proportion of variance in dorsal brain regions. Our findings demonstrate the potential for using task-relevant models to probe representational differences across visual streams.
% Using the large-scale fMRI Natural Scenes Dataset (NSD), we compared GRNN to self-supervised baseline models that have been shown to align well with ventral regions and found that GRNN accounted for a greater proportion of variance in dorsal brain regions than the baseline models, while the baselines accounted better for ventral brain regions.
%Our findings suggest that a 3D view prediction model aligns more closely with the dorsal visual stream than 2D augmentation invariant self-supervised models.
}
\begin{quote}
\small
\textbf{Keywords:} 
Visual Streams. DNN. Self-Supervision. fMRI. 
\end{quote}

The visual cortex has been traditionally organized into two processing streams~\shortcite{ungerleider1982two}, the ventral and dorsal\footnote{Following~\shortciteA{finzi2022deep}, we refer to the dorsal stream as the parietal stream to avoid confusion with the lateral stream.} (parietal) pathways, with a third lateral pathway being proposed recently~\shortcite{weiner2013neural, wurm2022two, pitcher2021evidence}. 
Deep neural networks (DNNs) trained for object recognition
% supervised for object recognition or self-supervised to encode 2D image statistics have
% proven
have been found to be highly predictive of the ventral visual stream processing~\shortcite{Yamins2014}.
%\shortciteA{Yamins2016} posit that common tasks across artificial and biological vision lead to representations that are well-aligned across the two domains.
However, it remains unclear whether DNNs for recognition are well suited for predicting non-ventral visual processing, in the lateral or parietal visual streams.

% \vspace{}ral earlier projects sought to identify DNNs that solved tasks better aligned with dorsal processing. For example,
DNNs optimized for egomotion estimation or action recognition may better predict neural responses in the parietal and lateral visual streams~\shortcite{mineault2021your, gucclu2017increasingly}. However, DNNs trained for action recognition do not appear to differentiate themselves from DNNs trained for object recognition in terms of predicting activity across the visual streams~\shortcite{finzi2022deep}. Recent experimental work 
%examining the role of the dorsal visual stream during recognition tasks 
has demonstrated evidence that the parietal stream plays a major role in global shape perception during object recognition, while the ventral stream may be more involved in local shape and texture encoding~\shortcite{ayzenberg2022dorsal}. Additionally, a well-established function of the parietal pathway is depth and 3-D shape perception~\shortcite{welchman2016human}, and it has been suggested that representations in these areas may arise from self-supervised predictive coding~\shortcite{jehee2006learning, raman2016predictive, bakhtiari2021functional}. 
% Additionally, the large amount of supervision used to train these models makes them less biologically plausible, as pointed out in previous literature [cite], while predictive coding theory has been widely proposed as a more biologically plausible training signal for the brain [cite].

\begin{figure}[ht]
\begin{center}
\includegraphics[width=9cm]{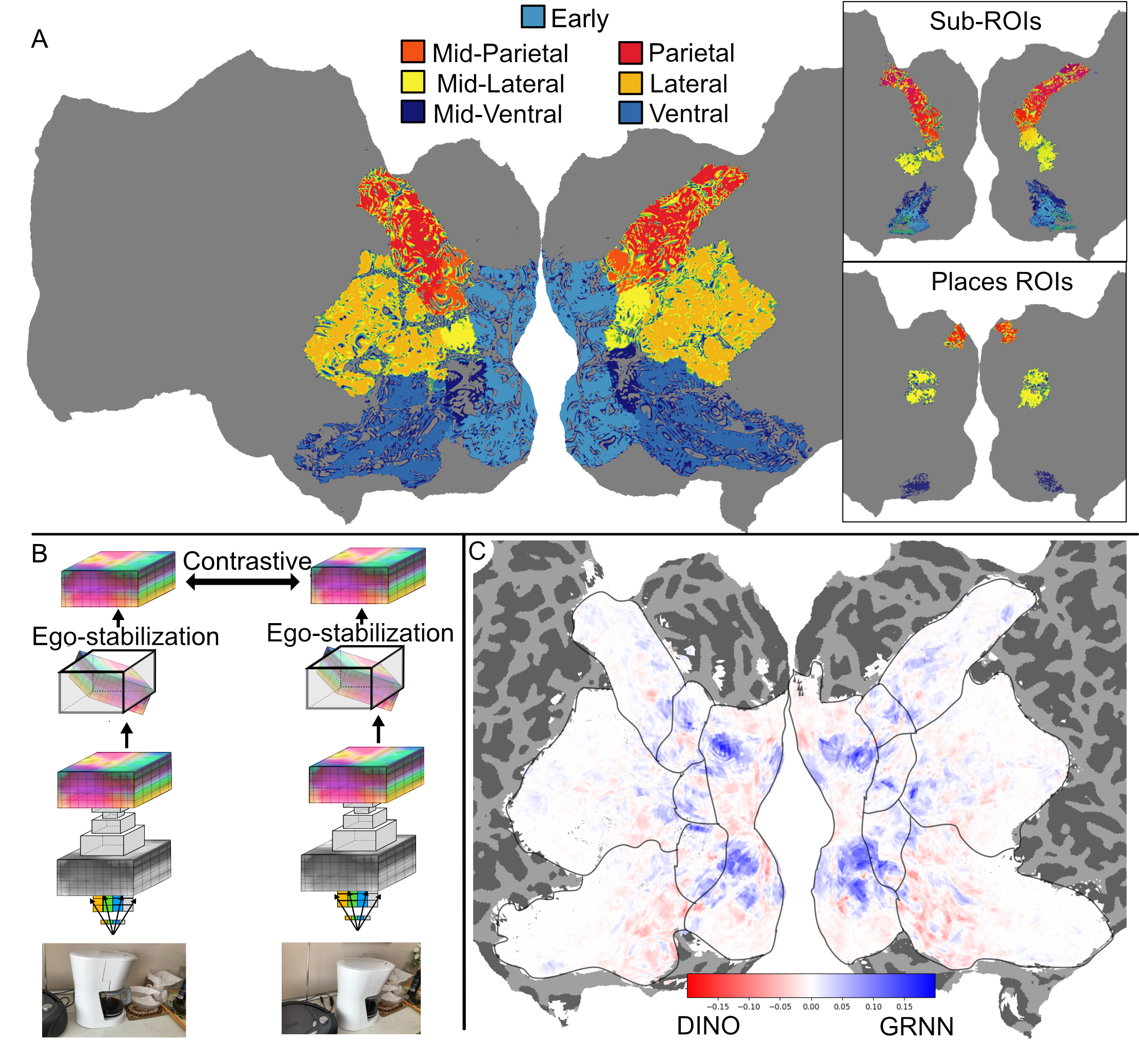}
\end{center}
\vspace*{-3mm}
\caption{\small \textbf{A.} ROI definitions on a flat map of the fsaverage cortex. \textbf{B.} GRNN training procedure. \textbf{C.} Difference (subtraction) between GRNN and DINO noise-corrected $R^2$'s for each voxel for example  S1. We observe higher $R^2$ for GRNN (darker blue) in high-level parietal regions, and conversely higher $R^2$ for DINO (darker red) in high-level ventral regions.} 
\vspace{-.4 cm}
\label{sample-figure}
\end{figure}

What kinds of neural networks might best account for neural processing in the parietal visual stream? We propose that the ``GRNN'' model from \shortciteA{tung2019learning} is a promising ``proxy model''~\shortcite{Leeds2013} for investigating computational constraints within the parietal pathway. GRNN learns spatially-aware 3D representations of visual inputs and is trained in a self-supervised manner to predict the complete 3D feature representation of a scene from one camera viewpoint, given input from another camera viewpoint. The model can ``fill in'' and predict features in the 3D feature map to represent the complete 3D geometry and shape of a scene from a partial 2.5D view. In this sense, the task solved by the GRNN model aligns well with several of the commonly proposed functional characteristics of the parietal stream.

We examined GRNN's ability to predict neural responses as measured by fMRI in response to viewing complex, natural scenes \shortcite{Allen2022}. As baseline models, we used self-supervised DNNs that were trained to maximize agreement between different augmentations of 2D images, which have previously been shown to be highly predictive of ventral visual stream~\shortcite{caron2021emerging, chen2020simple}. Consistent with our proposal, we found that the GRNN model was typically a better predictor of high-level parietal visual areas, while the self-supervised 2D models were typically better predictors of high-level ventral visual areas. These results demonstrate the potential for using task-relevant models aligned with hypotheses regarding brain function as a means for probing representational differences across visual streams.
%the underlying computations.
% using self-supervised 3D view-consistent DNNs as a strong candidate for the dorsal visual stream and suggest that models beyond DNNs optimized for 2D image statistics may better model the dorsal visual stream during image recognition.

\vspace{-.2 cm}
\section{Methods}
\vspace{-.2 cm}
\textbf{fMRI dataset.} 
% We used the Natural Scenes Dataset (NSD) (Allen et al., 2022), which consists of high-resolution fMRI responses to naturalistic scenes. NSD contains fMRI data from 8 screened subjects who each viewed 6234–10,000 scene images. We define X regions of interest... 
NSD contains measurements of 7T fMRI responses (1.8~mm, 1.6~s) from 8 participants who each viewed 9,000–10,000 distinct color natural scenes (22,000–30,000 trials). Participants fixated centrally and performed a long-term continuous image recognition task. The noise ceiling (NC) was estimated in each voxel as described in \shortciteA{Allen2022}. We only include voxels with NC $\geq$ 10\% variance and report noise-ceiling normalized prediction accuracy. \\
\textbf{Regions of Interest (ROIs).} 
% We used the ``streams'' anatomical atlas region of interest (ROI) definitions from NSD (Fig. 1a; also see Finzi et. al., 2022). These definitions are based on the fsaverage folding and noise ceiling results and comprise seven ROIs that span the parietal, lateral, and ventral visual streams. We used an early visual cortex (EVC) ROI, as well as intermediate and higher-level ROIs for each of the three proposed streams. We also examined several sub-regions within each of the streams, which were obtained from the Glasser et al. (2016) and Wang et al. (2015) atlases. We also examined three scene ROIs obtained by thresholding the category functional localizer, which include RSC, OPA, and PPA.\\
We used NSD's "streams" anatomical atlas to define seven ROIs that cover the parietal, lateral, and ventral visual streams (Fig. 1A; also see \shortciteA{finzi2022deep}). We also looked at sub-regions within each stream using \shortciteA{glasser2016multi} and \shortciteA{wang2015probabilistic} atlases. We also examined three scene ROIs (RSC, OPA, and PPA) obtained by thresholding the category functional localizer. \\
\textbf{GRNN training and inference.} 
We used the \shortciteA{fang2020move} dataset of RGB-D images ($n = 28345$) of indoor~\shortcite{replica19arxiv} and outdoor~\shortcite{Dosovitskiy17} scenes for our GRNN training. The self-supervised training procedure from \shortciteA{harley2019learning} was used, which utilizes a view-contrastive loss in feature space. This involves back-projecting an RGB image into a 3D voxel grid, deriving a 3D feature map, and pulling corresponding features together from egomotion-stabilized 3D feature maps (Fig. 1B). To extract GRNN representations for NSD images, we used a fixed camera field of view and estimated depth maps using MiDaS~\shortcite{ranftl2020towards}. \\
\textbf{Comparison models.} 
% We compare GRNN to two self-supervised DNNs that have demonstrated state-of-the-art performance in object recognition, even approaching that of supervised models for object recognition and ventral stream predictivity~\shortcite{zhuang2021unsupervised}. These models vary in both their self-supervised learning objective and neural architecture, and include DINO~\shortcite{caron2021emerging} (ViT-small backbone) and SimCLR~\shortcite{chen2020simple} (ResNet-50 backbone). To ensure a fair comparison with GRNN, we trained all models on the same dataset of indoor and outdoor scenes. The default hyperparameters were used for training each model.\\
We compared GRNN to two self-supervised DNNs that have shown exceptional performance in object recognition and ventral stream predictivity, even rivaling supervised models~\shortcite{zhuang2021unsupervised}. These models, DINO~\shortcite{caron2021emerging} (ViT-small backbone)  and SimCLR~SimCLR~\shortcite{chen2020simple} (ResNet-50 backbone), have different self-supervised learning objectives and neural architectures. We trained all models on the same dataset of indoor and outdoor scenes to ensure a fair comparison with GRNN.\\
% \textbf{Fitting to brain data.} We extracted the activations from the output of all layers of each model for each image in the NSD dataset. Performance was evaluated on a held-out test set (85 validation/15 test split) for each subject separately. We project the features into a lower dimensional subspace using PCA fit to the validation images, and retain the first 1000 components, following [cite]. For each subject, we fit the model features to each brain voxel independently via ridge regression, where each voxel’s regularization parameter via 7-fold cross-validation based on the prediction performance of the validation data. Model performance was evaluated on the test data using both Pearson’s correlation and coefficient of determination ($R^2$
% ). 
\textbf{Fitting to brain data.} 
We evaluated the performance of each model on a held-out test set using an 85:15 validation split for each subject separately. To reduce dimensionality, we used PCA to project the features into a lower dimensional subspace and retained the first 1000 components~(\shortciteA{schrimpf2018brain}). We fit the features of each layer to each brain voxel using ridge regression, and determined each voxel's regularization parameter through 7-fold cross-validation. We assessed model performance on the test data using Pearson's correlation and coefficient of determination ($R^2$), and reported the best fitting layer for each subject in each ROI.
\begin{figure}[ht]
\begin{center}
\includegraphics[width=7cm]{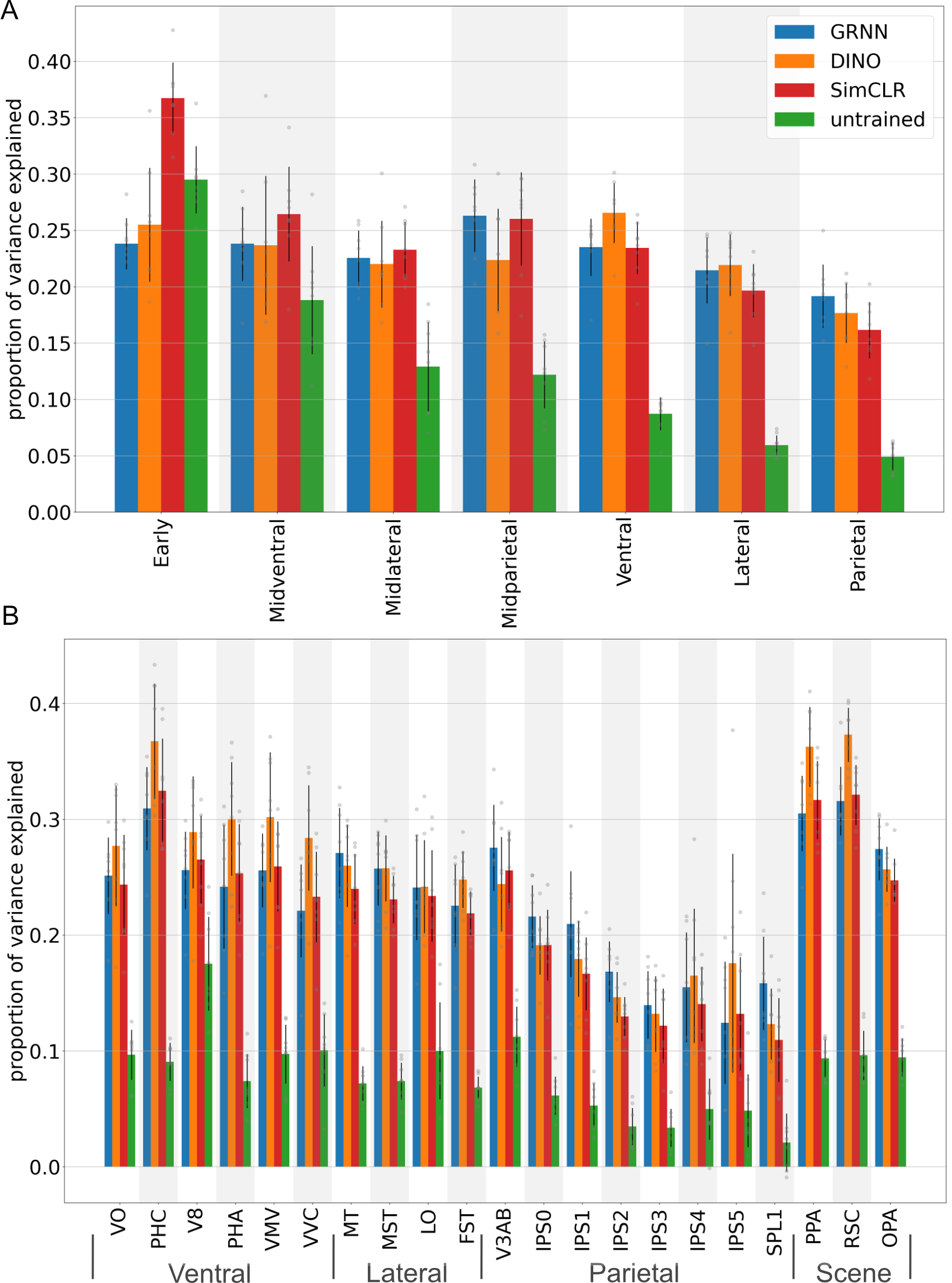}
\end{center}
\vspace*{-3mm}
\caption{\small \textbf{A.} noise-corrected $R^2$ of model fit for each ROI. Each dot represents a subject. Bars and error bars are mean and standard deviation across subjects (N=8), respectively. \textbf{B.} Same as A but for subregion ROIs.} 
\vspace*{-2mm}
\label{sample-figure}
\end{figure}

\vspace{-.4 cm}
\section{Results}
\vspace{-.2 cm}
We evaluated the performance of three models for predicting voxel responses to natural images. A representative subject (Fig. 1C) indicates that GRNN outperforms DINO in high-level parietal regions, while DINO performs better in high-level ventral regions. Prediction accuracy ($R^2$) within each stream across all subjects reveals that GRNN predicts high-level parietal regions better than DINO and SimCLR (GRNN$>$DINO mid-parietal $p=0.016$, high-parietal $p=0.038$; GRNN $>$ SimCLR, high-parietal $p=0.005$). DINO also performs better than GRNN in high-level ventral regions (paired $t$-test; $p=0.0052$). There was no significant difference between GRNN and SimCLR in high-level ventral (paired $t$-test; $p=0.98$) and no significant difference between models in the mid- and high-level lateral ROIs. 
% Further analysis on sub-ROIs within each stream showed that GRNN predicts voxel responses with significantly higher accuracy than DINO in V3AB (V3AB; paired t-test; $p=0.024$), and early IPS regions and SPL in DINO and SimCLR (IPS0, IPS1, IPS2, SPL; paired t-test; $p<0.05$), but we did not observe a significant difference between the models in higher-order regions of IPS (IPS3, IPS4, IPS5; paired t-test; $p>0.05$). In 4/6 high-level ventral regions examined, DINO had significantly higher predictivity compared to GRNN (PHC, PHA, VMV, VVC; paired t-test; $p<0.05$; VO, V8 no significant difference $p>0.05$). Our analysis also revealed that DINO and SimCLR significantly outperform GRNN in predicting voxel responses in scene ROIs located more ventrally (PPA, RSC; paired t-test; $p<0.05$), while GRNN performs better in predicting voxel responses in OPA, a scene ROI located in dorsal regions of the brain (OPA; paired t-test; $p<0.05$). In lateral sub-ROIs, we did not observe a substantial pattern between the models.
Paired $t$-tests on sub-ROIs within each stream revealed that GRNN predicts voxel responses with higher accuracy than DINO in V3AB ($p=0.021$), as well as early IPS regions and SPL compared to DINO (GRNN$>$DINO, IPS0; $p=0.017$, IPS1; $p=0.035$, IPS2; $p=0.039$, SPL; $0.0078$) and SimCLR (GRNN$>$SimCLR, IPS0; $p=0.018$, IPS1; $p=0.010$, IPS2; $p=0.013$, SPL; $0.015$). No difference between models was found in higher-order IPS (GRNN$>$DINO, IPS3; $p=0.22$, IPS4; $p=0.35$, IPS5; $p=0.058$). 
In high-level ventral regions DINO had significantly higher predictivity than GRNN in 4/6 regions examined (DINO$>$GRNN, PHC; $0.023$, PHA; $0.0019$, VMV; $0.047$, VVC; $p=0.00066$). In scene ROIs DINO significantly outperformed GRNN in two scene regions located more ventrally (DINO$>$GRNN, PPA; $p=0.0034$, RSC; $0.00091$), whereas GRNN performed better in predicting voxel responses in OPA, which is located more dorsally (GRNN$>$DINO $p=0.011$; GRNN$>$SimCLR $p=0.037$). 
% We did not observe any noteworthy pattern between the models in lateral sub-ROIs.
% , with DINO and SimCLR not significantly different in most cases, except GRNN significantly predicting MT and MST significantly greater than SimCLR (paired t-test; $p<0.05$), and DINO significantly 
%found GRNN to predict MT and MST signficantly greater than SimCLR, and DINO to predict to predict FST signficantly greater than SimCLR, but 
% found most  having no significant differences (paired t-test; $p>0.05$). 
% These findings indicate that a self-supervised model trained for 3D view prediction is a better predictive model of parietal areas compared to static 2D self-supervised models, while the reverse may be true for ventral regions as indicated by our results showing DINO outperforming GRNN in ventral regions. 
These results indicate that a self-supervised model trained for 3D view prediction performs better than models trained to capture augmentation-invariant 2D image statistics in predicting voxel responses in parietal areas. The opposite trend was found in ventral regions, particularly with DINO outperforming GRNN.

\vspace{-.2 cm}
\section{Discussion}
Our research indicates that a 3D view prediction model is better suited for predicting voxel responses in the parietal visual stream compared to 2D augmentation-invariant self-supervised models in a large-scale fMRI dataset of humans viewing natural images. However, more research is necessary to better understand the observed differences
% specific aspects of the models that cause the observed differences 
and to explore the impact of training and fMRI datasets on model alignment.

\clearpage
\section{Acknowledgments}
This material is based upon work supported by National Science Foundation grants GRF DGE1745016 \& DGE2140739 (GS), a DARPA Young Investigator Award, a NSF CAREER award, an AFOSR Young Investigator Award, and DARPA Machine Common Sense. Any opinions, findings and conclusions or recommendations expressed in this material are those of the authors and do not necessarily reflect the views of the United States Army, the National Science Foundation, or the United States Air Force.

% \section{References Instructions}

% Follow the APA Publication Manual for citation format, both within the
% text and in the reference list, with the following exceptions: (a) do
% not cite the page numbers of any book, including chapters in edited
% volumes; (b) use the same format for unpublished references as for
% published ones. Alphabetize references by the surnames of the authors,
% with single author entries preceding multiple author entries. Order
% references by the same authors by the year of publication, with the
% earliest first.

% Use a first level section heading, ``{\bf References}'', as shown
% below. Use a hanging indent style, with the first line of the
% reference flush against the left margin and subsequent lines indented
% by 1/8~inch. Below are example references for a conference paper, book
% chapter, journal article, dissertation, book, technical report, and
% edited volume, respectively.

\nocite{ChalnickBillman1988a}
\nocite{Feigenbaum1963a}
\nocite{Hill1983a}
\nocite{OhlssonLangley1985a}
% \nocite{Lewis1978a}
\nocite{Matlock2001}
\nocite{NewellSimon1972a}
\nocite{ShragerLangley1990a}

\bibliographystyle{apacite}

\setlength{\bibleftmargin}{.125in}
\setlength{\bibindent}{-\bibleftmargin}

\bibliography{refs}

\end{document}